\pdfoutput=1
\documentclass[11pt]{article}
\usepackage[preprint]{acl}
\usepackage{times}
\usepackage{latexsym}
\usepackage[T1]{fontenc}
\usepackage[utf8]{inputenc}
\usepackage{microtype}
\usepackage{inconsolata}
\usepackage{graphicx}
\usepackage{booktabs}
\usepackage{amsmath}
\usepackage{amssymb}
\usepackage{nicefrac}
\usepackage{xcolor}
\usepackage{colortbl}
\usepackage{multirow}
\usepackage{makecell}

\newcommand{\leto}{Leto}
\definecolor{paperfill}{HTML}{C6D7F4}
\colorlet{oursbg}{paperfill!100!white}

\title{Leto: Utility-Centric Agent Memory}
\author{%
  Shuguang Chen \\
  DigitalOcean \\
  \texttt{schen@digitalocean.com}
\And
  Salman Paracha \\
  DigitalOcean \\
  \texttt{salmanparacha@digitalocean.com}
}

\begin{document}
\maketitle
\begin{abstract}
Large language model (LLM) agents rely on external memory to retain and reuse information across interactions.
Existing memory systems treat memory as a record of the past and evaluate it by fidelity to the conversation.
However, the value of memory lies in whether it improves the agent's response.
A memory system decides what to memorize and what to serve before that response exists, and fidelity checks neither decision against it.
Faithful memory can therefore fail to improve the response.
We instead view memory as a prediction problem and present \textbf{\emph{\leto{}}}, a memory system for LLM agents that focuses on the utility of memory to the future rather than on its fidelity to the past.
Because future requests are unknown, \emph{\leto{}} preserves access to conversation details through memory indexing, allowing information to remain available through its sources.
When a request arrives, memory replay draws on this index to anticipate and supply the context the agent needs beyond what the request explicitly states.
The utility of the served context remains uncertain until the response is generated, motivating a utility loop that uses response outcomes to refine future memory selection.
We evaluate our system against six memory systems on five benchmarks across three scenarios.
In personalization, it outperforms the strongest compared memory system by 16.3 percentage points in accuracy on PrefEval and by 3.1 points in task completion on PERMA.
For long-horizon tasks, it achieves the best overall performance on MemoryArena averaged over five environments with a gain that increases with the chain depth.
In long-context understanding, it is on par with retrieval over the raw history while serving fewer context tokens.
\end{abstract}

\section{Introduction}
\label{sec:intro}

An LLM agent that serves the same user or works on the same task across sessions requires memory of what earlier sessions established, such as a preference the user expressed weeks ago or the state of an unfinished task. Existing memory systems extract facts or generate summaries from each session, then retrieve the memory most similar to an incoming request and insert it into the context~\citep{chhikara2025mem0,rasmussen2025zep,latimer2025hindsight}. Some systems additionally reflect on the stored memory in the background to form higher-level notes or user profiles~\citep{park2023generative,zhong2024memorybank,xu2025amem}. Both the construction and the evaluation of these systems are oriented toward the past. Their evaluation measures fidelity (i.e., how completely the memory covers the conversation and how often the retrieved memory contains the gold evidence)~\citep{maharana2024evaluating,wu2025longmemeval,liu2026perma}.

\begin{figure*}[t]
  \centering
  \includegraphics[width=\linewidth]{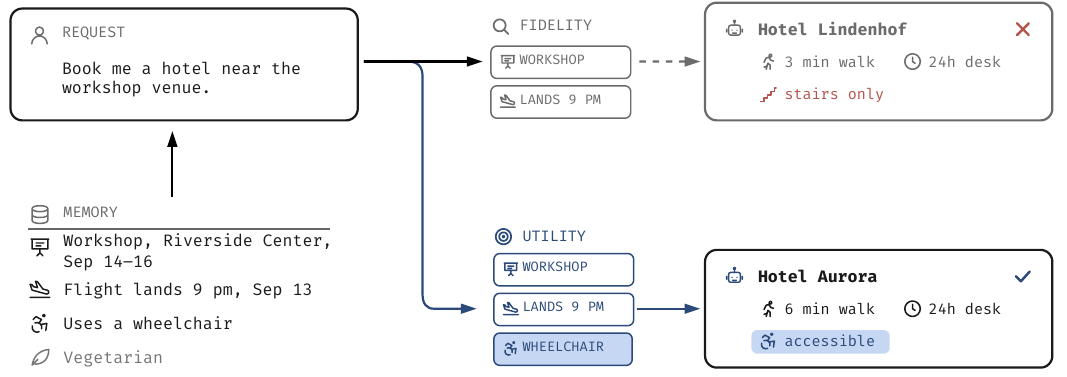}
  \caption{An example of a request served from the same memory store under different criteria. Selection by fidelity serves the memory that matches the request; selection by utility serves the memory the response needs. Hotel names and details are fictional.}
  \label{fig:fidelity}
\end{figure*}

However, fidelity to the past does not by itself yield a better response. An agent needs the relevant context in its input to produce a good response. A memory system supplies this context through two decisions: what to memorize when a conversation ends and what to serve when a request arrives. Existing systems assume that fidelity at each stage preserves the context needed for future responses. This assumption leaves two gaps: 1) The memorization gap arises because what is memorized determines what can later be served, yet faithful memory does not necessarily preserve what a future request will require. Existing systems keep summaries that omit details whose relevance appears later~\citep{wu2025longmemeval,zou2026demem}, or extract facts judged by coverage of the conversation, not by what a future request requires~\citep{chhikara2025mem0,latimer2025hindsight}; 2) The serving gap arises because memory is selected by how closely it matches the request, which may not reflect what the response requires. Existing systems rank memory by similarity to the request, yet users rarely restate standing preferences and memory that would change the response may not rank highly~\citep{salemi2024lamp,zhao2025prefeval}. Fidelity measures neither gap because it assesses how well memory represents the conversation, not whether the served context meets the needs of the response.

Both gaps arise because memory decisions must anticipate the context a response will need before their effects can be evaluated. We therefore view memory as a prediction problem. At write time, the system must predict which information will matter for future requests. At serving time, it must predict which memory the response needs. The request provides evidence about these needs but does not fully specify them. The accuracy of these predictions remains unknown until the response is generated. Both should therefore be evaluated against the observed outcome, which can then be used to refine future predictions. The value of memory thus lies in the response it enables, which we call its utility. Figure~\ref{fig:fidelity} illustrates the difference between the two criteria on one request served from the same store. Selection by fidelity serves the memory that matches the request and nothing more. The response satisfies the request as stated, yet the user cannot use the hotel it books. Selection by utility also serves the memory that the user uses a wheelchair, which the request does not mention. The response then books a hotel the user can use. The two selections differ by a single memory, which decides the outcome.

We build on this view and present \textbf{\emph{\leto{}}}, a memory system that manages memory by its predicted utility. To address the memorization gap, \emph{\leto{}} preserves access to conversation details through memory indexing so that information remains available even when its utility is not apparent at write time. To address the serving gap, memory replay draws on this index ahead of the request to anticipate the context the agent needs beyond what the request states. This context is served together with the memory retrieved for the request. The response then provides evidence of the utility of what was served, and a utility loop feeds this evidence back to refine subsequent memory selection.

We evaluate \emph{\leto{}} against six memory systems on five benchmarks spanning three scenarios: personalization, long-horizon tasks, and long-context understanding. \emph{\leto{}} achieves the best overall performance among the compared memory systems in all three scenarios. In personalization, it outperforms the strongest compared memory system by 16.3 percentage points in accuracy on PrefEval and by 3.1 points in task completion on PERMA. For long-horizon tasks, it achieves the best overall performance on MemoryArena averaged over five environments. Its gain over an agent without memory increases with the chain depth. For long-context understanding, it is competitive with retrieval over the raw history while serving fewer context tokens.

\section{Related Work}
\label{sec:related}

External memory systems build memory of past interactions and serve parts of it to later ones. They extract facts or write summaries and reconcile them as interactions accumulate~\citep{packer2023memgpt,chhikara2025mem0,zhong2024memorybank}. They organize the store as temporal graphs or linked notes or associative indexes~\citep{rasmussen2025zep,xu2025amem,gutierrez2024hipporag}. They add editable memory blocks and reflection over the store alongside several retrieval mechanisms~\citep{packer2023memgpt,park2023generative,latimer2025hindsight}. They are also evaluated against the past. The write side is judged by how completely and accurately memory captures the conversation, the read side by how well it ranks memory against the request. The benchmarks measure memory the same way~\citep{wu2025longmemeval,maharana2024evaluating,liu2026perma}.

Several lines of work move one of the two decisions toward the future. On what to memorize, some index retained text by decontextualized claims or extracted entities~\citep{choi2021decontextualization,chen2024densex,gutierrez2024hipporag,wu2025longmemeval}. Others keep what a later decision is expected to need~\citep{zou2026demem}. On what to serve, some place standing memory in the context or revise memory between interactions before any request arrives~\citep{packer2023memgpt,zhong2024memorybank}. Memory-R1~\citep{yan2025memoryr} learns memory operations from downstream task reward. None carries on the memory itself an estimate of its utility that feedback updates. \emph{\leto{}} moves both decisions toward the future and maintains such an estimate for every memory.

Whether memory helps is a question about outcomes. PrefEval~\citep{zhao2025prefeval} and PERMA measure whether an earlier preference is followed and a personalized task completed. MemoryArena~\citep{he2026memoryarena} measures progress on interdependent tasks. LoCoMo and LongMemEval measure answers to questions over long histories. Each reports a system's level. The utility of its memory is the level's difference from the same agent without memory on the same items, which can be negative and which no level shows.

\section{Problem Formulation}
\label{sec:problem_formulation}

\begin{figure*}[t]
  \centering
  \includegraphics[width=\linewidth]{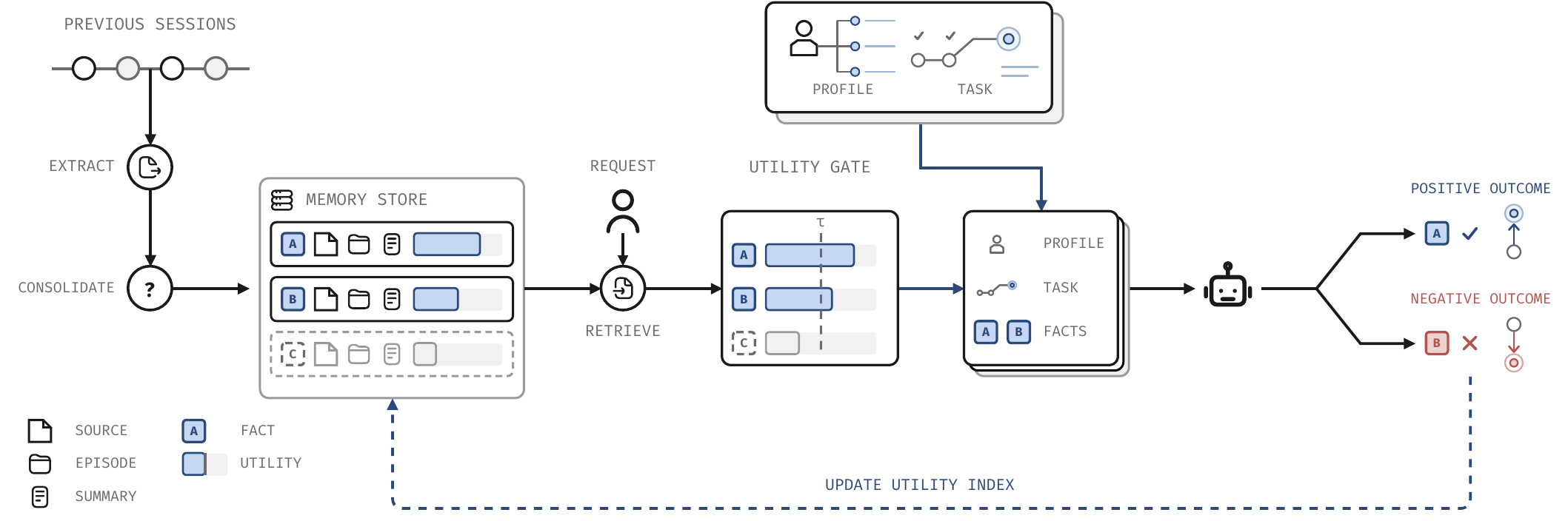}
  \caption{\emph{\leto{}} workflow. Memory indexing stores atomic facts as keys to their retained source. Memory replay composes the profile and task state that give context across requests. The utility loop gives each new memory a forecast $u_0(m)$ and serves retrieved memory whose index satisfies $u \geq \tau$, then updates the index from feedback.}
  \label{fig:overview}
\end{figure*}

An external memory system supports an agent across timestamped sessions by deciding what to memorize and what to serve. At \emph{write time}, it decides what to memorize when a session ends. At \emph{serving time}, it selects memory $M$ from the store to place in the agent's context when a request $x$ arrives. The fidelity view judges the write decision by how completely and accurately the store represents past sessions, and judges the serving decision by how closely the served memory matches $x$. Both criteria can be assessed using information available at decision time, but neither establishes whether the context needed for the right response is supplied. The \emph{memorization gap} consists of required information from past sessions that the store no longer holds. The \emph{serving gap} consists of required information that the store holds but that the served memory omits.

We instead define the objective by task outcomes. Let $y(x,M)\in\{0,1\}$ denote the graded task outcome for request $x$ when memory $M$ is supplied to the agent. We define the utility of the served memory as
\begin{equation}
U(x,M) = y(x,M)-y(x,\varnothing),
\label{eq:utility-def}
\end{equation}
where $y(x,\varnothing)$ is the outcome without memory. This objective captures whether the served memory improves or hurts task performance, or leaves it unchanged. It connects both memory decisions to task performance. The write decision determines what information remains available, and the serving decision determines what reaches the agent.

Utility depends on both the request and how the agent uses the served memory. Relevant information may provide no benefit if it is redundant or unused, and may hurt performance if it misleads the agent. The system must therefore anticipate what the served memory contributes to the response when deciding what to serve. Future requests are unknown at write time, so memory utility must be forecast across them. During subsequent interactions, feedback provides proxy signals of memory utility that update the utility index for the serving decisions that follow.
\section{Methodology}
\label{sec:methodology}

\emph{\leto{}} manages memory through three mechanisms, memory indexing and memory replay followed by the utility loop. Figure~\ref{fig:overview} summarizes this workflow.

\subsection{Memory Indexing}
\label{sec:write}

Because future requests are unknown at write time, a memory system may retain information that is never needed or discard information that is. The costs of these two errors are asymmetric. Unhelpful information costs storage, while a discarded detail degrades the performance on every later request that depends on it. \emph{\leto{}} therefore retains every conversation and indexes it to find what a request needs.

\paragraph{Write time.}
Indexing the conversation by its raw messages would make retrieval imprecise, because a message can cover several topics and its meaning often depends on its context. \emph{\leto{}} therefore indexes the conversation by atomic facts~\citep{choi2021decontextualization,chen2024densex}. An atomic fact states a single claim in self-contained form. This allows it to be matched to a request precisely and compared with other facts on the same subject. Existing memory systems commonly serve the extracted facts themselves as the memory. \emph{\leto{}} instead uses each fact as a key whose value is its source, the span of adjacent messages that the fact was extracted from. A source reaches the agent only when a fact matching the request points to it, so the context carries less noise than retrieval over raw chunks. Information omitted from a fact also remains accessible through its source, which preserves the original wording and qualifications. New facts are consolidated with the existing ones. A fact that revises an earlier one supersedes it without deleting it. The index can thus distinguish what currently holds from what once held. Each session also receives a short summary.

\paragraph{Serving time.}
When a request arrives, \emph{\leto{}} retrieves the relevant facts with their sources. The utility loop (Section~\ref{sec:utility}) then decides which are served within a token budget, alongside the standing context from memory replay (Section~\ref{sec:serve}).

\subsection{Memory Replay}
\label{sec:serve}

Retrieval can miss relevant preferences and constraints that a request leaves unstated. \emph{\leto{}} therefore replays stored memory between conversations. It re-reads what the store holds and composes the standing context that the next conversation starts with. The standing context consists of a user profile and a task state if a task is declared. Both are composed before any request arrives.

\paragraph{Profile rewrite.}
Profile rewrite maintains a user profile that carries what holds for the user regardless of the request, such as standing preferences and constraints or instructions on how to act. After new sessions are indexed, \emph{\leto{}} rewrites the profile from the current facts about the user and the session summaries. The rewrite keeps the profile concise by merging related facts into one statement and leaving out superseded preferences.

\paragraph{Task rendering.}
A later step of a long-horizon task depends on what earlier sessions established, which the request rarely restates. Task rendering supplies it as a task state at the start of every conversation on the task. The first part of the task state shows where the task stands through its current facts, grouped by entity with each attribute and its current value. The second part shows how the task got there through summaries of its prior sessions in order.

\subsection{Utility Loop}
\label{sec:utility}

Whether a memory will improve a response cannot be observed at write time. \emph{\leto{}} therefore forecasts a utility index $u(m)\in[0,1]$ for each memory and revises it with the outcome observed each time the memory is served. This cycle forms the utility loop.

\paragraph{Utility forecast.}
Each new memory receives an initial utility forecast $u_0(m)$ as its utility index. A low forecast does not delete the memory. The fact and its source remain in the store. Retention and serving are separate decisions, which allows serving priorities to change as feedback accumulates.

\paragraph{Outcome observations.}
The forecast is made before a memory is used, and whether the memory helps becomes evident only after it is served. \emph{\leto{}} therefore observes the outcome of each served memory to obtain evidence of its utility. We use the interaction signals from \citet{chen2026signals} as the observations, which are detected in the conversation by lexical rules. Each signal marks a positive outcome (e.g., a confirmation or a use of the memory) or a negative outcome (e.g., a correction) for a memory, encoded as $1$ and $0$, respectively. When a graded task outcome is available, it can also be recorded for the memory served during the task. These signals provide evidence of memory utility without directly measuring an individual memory's contribution to the outcome.

\paragraph{Utility update.}
Given observations $o_1,\ldots,o_n$ and the initial forecast weighted as $k$ observations, the index is
\begin{equation}
  u(m) = \frac{k\,u_0(m)+\sum_{i=1}^{n}o_i}{k+n}.
  \label{eq:utility}
\end{equation}
Here, $k>0$ controls the weight of the initial forecast relative to feedback. The updated index then closes the loop. A retrieved memory passes the utility gate only if its index meets a threshold $\tau$. When the retrieved memory exceeds the context budget, memory with a higher utility index is prioritized. The utility index is a heuristic serving priority, not an estimate of Equation~\ref{eq:utility-def}. It combines observable signals to guide memory selection, while paired evaluation measures whether the system as a whole improves task outcomes.

\section{Experiments}
\label{sec:experiments}

We evaluate \emph{\leto{}} across three scenarios: personalization, long-horizon tasks, and long-context understanding. These scenarios differ in what information is needed from earlier interactions and how explicitly the current request identifies it.

\subsection{Experimental Setup}
\label{sec:exp-setup}

\paragraph{Benchmarks.}
We use five benchmarks grouped into three scenarios: PERMA~\citep{liu2026perma} and PrefEval~\citep{zhao2025prefeval} for \emph{personalization}; MemoryArena~\citep{he2026memoryarena} for \emph{long-horizon tasks}; and LoCoMo~\citep{maharana2024evaluating} and LongMemEval\textsubscript{S}~\citep{wu2025longmemeval} for \emph{long-context understanding}. Table~\ref{tab:benchmarks} in Appendix~\ref{app:benchmarks} summarizes the sample counts and evaluation metrics as well as each memory store's scope.

\paragraph{Compared methods.}
We compare \emph{\leto{}} with six memory systems: \emph{Mem0}~\citep{chhikara2025mem0}, \emph{LangMem}\footnote{\url{https://github.com/langchain-ai/langmem}}, \emph{Hindsight}~\citep{latimer2025hindsight}, \emph{Letta}~\citep{packer2023memgpt}, \emph{Zep}~\citep{rasmussen2025zep}, and \emph{Memobase}\footnote{\url{https://github.com/memodb-io/memobase}}. Each system uses the configuration provided by the benchmark when available and follows its own documented setup otherwise. We also include a \emph{RAG} baseline that retains the complete history and retrieves raw conversation chunks by embedding similarity to the current request~\citep{lewis2020rag}.

We establish two further baselines: \emph{No Memory} supplies no memory from earlier sessions, and \emph{Standalone} is the benchmark's own agent given the complete history in its context. Appendix~\ref{app:rates} reports the share of units each system improves and degrades against \emph{No Memory}.

\subsection{Scenario 1: Personalization}
\label{sec:exp-personalization}

Personalization requires an agent to apply information from earlier interactions even when the current request leaves it unstated. PrefEval focuses on preference following, while PERMA evaluates personalization across a user's interaction history.

\subsubsection{PrefEval}
\label{sec:exp-prefeval}

\paragraph{Main results.}
PrefEval evaluates whether an agent follows a preference stated earlier in the conversation when it answers the request. Following the benchmark's evaluation protocol, we report preference-following accuracy and error categories in Table~\ref{tab:personalization}. \emph{\leto{}} achieves 60.5\% accuracy, exceeding the strongest compared system \emph{Hindsight} by 16.3 percentage points. \emph{RAG} and \emph{Standalone} reach 13.1\% and 3.3\%, respectively. \emph{\leto{}} also reduces preference-unaware violations to 19.1\%, compared with 33.8\% under \emph{Hindsight}.

\begin{table*}[t]
  \centering
  \footnotesize
  \setlength{\tabcolsep}{3.5pt}
  \begin{tabular}{l|ccccc|ccc|ccc}
    \toprule
    \multicolumn{1}{c|}{} & \multicolumn{5}{c|}{\textbf{PrefEval}} & \multicolumn{3}{c|}{\textbf{PERMA C-S}} & \multicolumn{3}{c}{\textbf{PERMA C-M}} \\
    \cmidrule[\heavyrulewidth](lr){2-6} \cmidrule[\heavyrulewidth](lr){7-9} \cmidrule[\heavyrulewidth](lr){10-12}
    \multicolumn{1}{l|}{\textbf{System}} & \multicolumn{1}{c}{\textbf{Unaware}} & \multicolumn{1}{c}{\textbf{Halluc.}} & \multicolumn{1}{c}{\textbf{Incons.}} & \multicolumn{1}{c}{\textbf{Unhelp.}} & \multicolumn{1}{c|}{\textbf{Acc.}} & \multicolumn{1}{c}{\textbf{MCQ}} & \multicolumn{1}{c}{\textbf{Comp.}} & \multicolumn{1}{c|}{\textbf{MS}} & \multicolumn{1}{c}{\textbf{MCQ}} & \multicolumn{1}{c}{\textbf{Comp.}} & \multicolumn{1}{c}{\textbf{MS}} \\
    \midrule[\heavyrulewidth]
    \multicolumn{12}{l}{\textbf{\emph{No History}}} \\
    \midrule
    No Memory & 74.00 & 24.20 & 0.00 & 0.40 & 1.40 & 53.66 & 74.23 & 1.00 & 56.69 & 60.51 & 1.00 \\
    \midrule
    \multicolumn{12}{l}{\textbf{\emph{Full History}}} \\
    \midrule
    Standalone (gpt-4o-mini) & 65.40 & 31.00 & 0.20 & 0.10 & 3.30 & 78.01 & 81.56 & 1.17 & 71.34 & 63.69 & 1.62 \\
    RAG & 54.20 & 31.70 & 1.00 & 0.00 & 13.10 & 70.45 & 83.69 & 1.87 & 68.15 & 61.78 & 1.73 \\
    \midrule
    \multicolumn{12}{l}{\textbf{\emph{Compared Memory Systems}}} \\
    \midrule
    Mem0 & 45.70 & \textbf{18.80} & 1.80 & 0.20 & 33.50 & 69.98 & 80.38 & 1.85 & 67.52 & 62.42 & 1.78 \\
    LangMem & 32.10 & 25.40 & 1.60 & 0.20 & 40.70 & 74.70 & 81.56 & 1.80 & 67.52 & 62.42 & 1.84 \\
    Hindsight & 33.80 & 19.90 & 1.90 & 0.20 & 44.20 & 69.03 & 80.14 & \textbf{2.06} & 66.24 & 59.87 & 1.90 \\
    Letta & 47.00 & 34.10 & \textbf{0.10} & 0.20 & 18.60 & 56.50 & 74.47 & \textbf{2.06} & 54.78 & 54.78 & \textbf{2.28} \\
    Zep & 68.10 & 20.90 & 0.60 & 1.00 & 9.40 & 68.32 & 78.49 & 1.57 & 58.60 & 63.69 & 1.46 \\
    Memobase & 45.80 & 24.30 & 1.50 & 0.20 & 28.20 & 73.29 & \textbf{81.80} & 1.78 & 67.52 & 64.97 & 1.73 \\
    \midrule
    \multicolumn{12}{l}{\textbf{\emph{Our System}}} \\
    \midrule
    \rowcolor{oursbg} \leto{} & \textbf{19.10} & \textbf{18.80} & 1.60 & \textbf{0.00} & \textbf{60.50} & \textbf{79.43} & 80.38 & 1.96 & \textbf{71.97} & \textbf{72.61} & 2.15 \\
    \bottomrule
  \end{tabular}
  \caption{Personalization results. PrefEval after ten filler turns (\%): four error types and accuracy (Acc.). Unaware: preference-unaware violation; Halluc.: hallucination; Incons.: inconsistency; Unhelp.: unhelpful response. Accuracy: the share of personalized responses, which satisfy the preference without violating or inventing one. PERMA type-3 (\%): C-S / C-M: clean single- / multi-domain; Comp.: completion; MS: memory score (1--4), PERMA's fidelity measure of the served context against the gold preferences. Bold: best value among the memory systems, the lowest for the four error types.}
  \label{tab:personalization}
\end{table*}

\begin{figure*}[t]
  \centering
  \includegraphics[width=\linewidth]{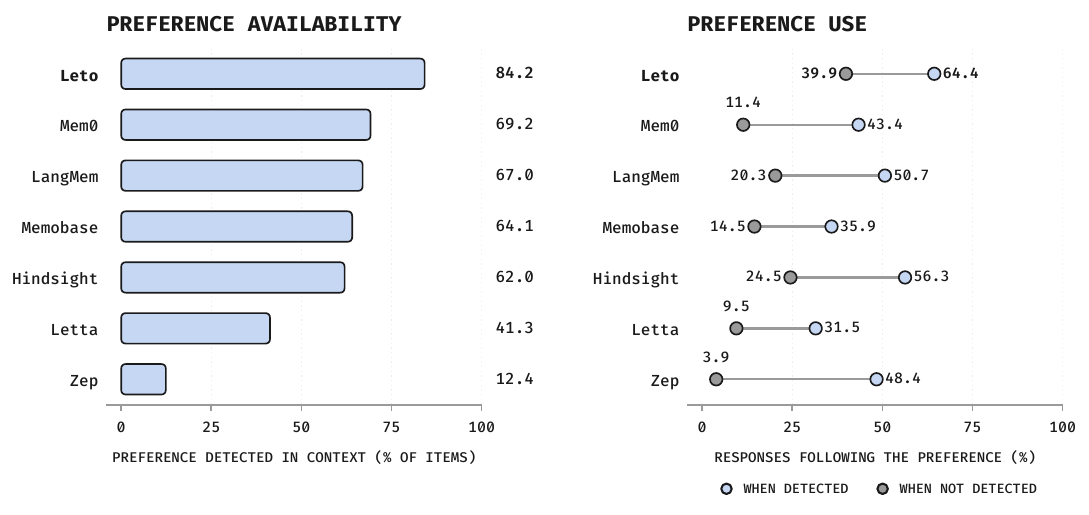}
  \caption{Preference availability and use on PrefEval. A preference counts as detected when at least 40\% of its content words occur in the served context. Blue and gray circles show preference-following rates with and without detection, respectively.}
  \label{fig:prefeval}
\end{figure*}

\paragraph{Preference availability and use.}
Accuracy on PrefEval depends on whether the preference reaches the agent's context and on whether the agent then follows it. Figure~\ref{fig:prefeval} separates the two, using lexical detection as a proxy for availability. A preference is detected when at least 40\% of its content words occur in the served context. The preference is detected in 84\% of \emph{\leto{}}'s contexts, compared with 12\%--69\% for the other memory systems. The agent follows a detected preference in 64\% of cases under \emph{\leto{}}, compared with 32\%--56\% under the other memory systems. The agent follows the preference more often when it is detected than when it is not, by 21--45 percentage points across the memory systems. Availability therefore matters, but it does not ensure use. \emph{Standalone} illustrates the difference. The preference is in its context inside the full history in every item, yet the agent follows it in 3.3\% of them. \emph{\leto{}} serves the preference in a short standing user profile written without reading the request and placed in every item, so it reaches the agent whether or not the request resembles it.

\begin{table*}[t]
  \centering
  \footnotesize
  \setlength{\tabcolsep}{3.5pt}
  \begin{tabular}{l|ccccc|cc}
    \toprule
    \multicolumn{1}{c|}{} & \multicolumn{5}{c|}{\textbf{Environment}} & \multicolumn{2}{c}{\textbf{Overall}} \\
    \cmidrule[\heavyrulewidth](lr){2-6} \cmidrule[\heavyrulewidth](lr){7-8}
    \multicolumn{1}{l|}{\textbf{System}} & \multicolumn{1}{c}{\textbf{Shopping}} & \multicolumn{1}{c}{\textbf{Travel}} & \multicolumn{1}{c}{\textbf{Search}} & \multicolumn{1}{c}{\textbf{Math}} & \multicolumn{1}{c|}{\textbf{Physics}} & \multicolumn{1}{c}{\textbf{PS}} & \multicolumn{1}{c}{\textbf{SR}} \\
    \midrule[\heavyrulewidth]
    \multicolumn{8}{l}{\textbf{\emph{No History}}} \\
    \midrule
    No Memory & 44.24 / 0.00 & 0.00 / 0.00 & 3.26 / 8.14 & 24.81 / 15.00 & 44.75 / 30.00 & 23.41 & 10.63 \\
    \midrule
    \multicolumn{8}{l}{\textbf{\emph{Full History}}} \\
    \midrule
    Standalone (gpt-5-mini) & 53.94 / 0.67 & 2.30 / 0.00 & 3.03 / 3.17 & 31.07 / 20.00 & 61.39 / 50.00 & 30.34 & 14.77 \\
    RAG & 49.69 / 0.00 & 1.50 / 0.00 & 5.85 / 13.57 & 33.07 / 22.50 & 57.85 / 45.00 & 29.59 & 16.21 \\
    \midrule
    \multicolumn{8}{l}{\textbf{\emph{Compared Memory Systems}}} \\
    \midrule
    Mem0 & 41.56 / 0.00 & 0.00 / 0.00 & 4.84 / 11.31 & 28.97 / 17.50 & 59.72 / 45.00 & 27.02 & 14.76 \\
    LangMem & 50.56 / 0.67 & 0.00 / 0.00 & 5.03 / 13.57 & 25.51 / 12.50 & 43.22 / 20.00 & 24.86 & 9.35 \\
    Hindsight & 43.05 / 0.00 & 0.00 / 0.00 & 4.19 / 9.95 & 30.31 / 17.50 & 54.26 / 30.00 & 26.36 & 11.49 \\
    Letta & \textbf{54.78 / 0.67} & 0.00 / 0.00 & 5.11 / 16.74 & 31.71 / 17.50 & \textbf{64.58 / 55.00} & 31.24 & 17.98 \\
    Zep & 46.63 / 0.00 & 0.00 / 0.00 & 6.99 / 21.27 & 27.27 / 12.50 & 44.47 / 20.00 & 25.07 & 10.75 \\
    Memobase & 48.55 / 0.00 & 0.00 / 0.00 & 5.21 / 14.03 & 23.92 / 0.00 & 47.18 / 20.00 & 24.97 & 6.81 \\
    \midrule
    \multicolumn{8}{l}{\textbf{\emph{Our System}}} \\
    \midrule
    \rowcolor{oursbg} \leto{} & 53.22 / 0.00 & \textbf{2.73 / 0.37} & \textbf{7.97 / 21.27} & \textbf{31.77 / 20.00} & 64.35 / 50.00 & \textbf{32.01} & \textbf{18.33} \\
    \bottomrule
  \end{tabular}
  \caption{MemoryArena progress score (PS) and success rate (SR) in percent, as PS / SR per environment. Overall columns average the five environments equally. SR measures full-chain completion except in search, where it measures final-question success. Bold: best value among the memory systems.}
  \label{tab:arena}
\end{table*}

\subsubsection{PERMA}
\label{sec:exp-perma}

\paragraph{Main results.}
We follow PERMA's type-3 evaluation, reporting multiple-choice (MCQ) accuracy and interactive task completion under the clean single-domain (C-S) and clean multi-domain (C-M) conditions. As shown in Table~\ref{tab:personalization}, \emph{\leto{}} achieves the highest MCQ accuracy in both conditions. Its clearest advantage is in C-M completion, where it exceeds the strongest compared system \emph{Memobase} by 7.64 percentage points. This advantage does not extend to C-S completion, where \emph{RAG} performs best. \emph{\leto{}} completes 76.5\% of tasks averaged over the two conditions, 3.1 points above \emph{Memobase}. Organizing the history matters most when a task draws on preferences from multiple domains: \emph{\leto{}}'s standing user profile gathers the user's preferences across domains into one block placed for every task, whereas retrieval by the request returns the domain the request names.

\paragraph{Fidelity and task performance.}
PERMA's memory score is a fidelity measure of the served context against the gold preferences and does not consistently track task performance. \emph{Letta} has the highest memory score in the C-M condition but the lowest MCQ accuracy and completion rate. A stronger fidelity score therefore does not necessarily translate into better personalization.

\subsection{Scenario 2: Long-Horizon Tasks}
\label{sec:exp-arena}

Long-horizon tasks require an agent to carry information from earlier interactions into later decisions. MemoryArena evaluates this ability through interdependent subtasks across shopping, travel, search, math, and physics.

\paragraph{Main results.}
Table~\ref{tab:arena} reports progress score (PS) and success rate (SR), averaged equally across the five environments. PS measures progress within a task chain, while SR measures the proportion of completed chains. \emph{\leto{}} achieves the highest overall PS (32.01\%) and SR (18.33\%) among all compared methods. It is also the only memory system whose useful rate exceeds its harmful rate in every environment (Appendix~\ref{app:rates}).

\begin{figure}[t]
\centering
\includegraphics[width=\columnwidth]{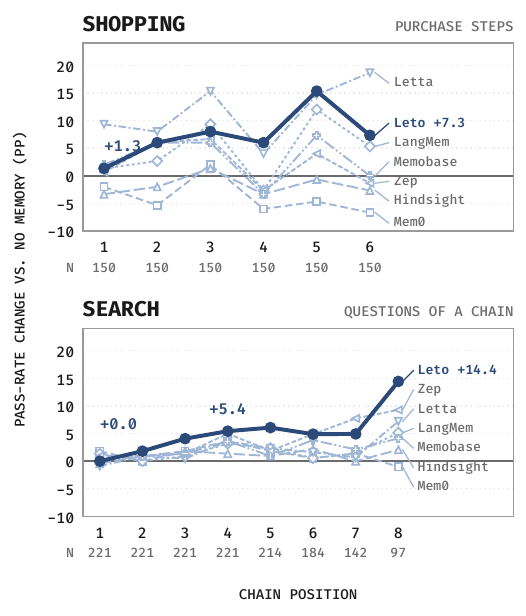}
\caption{Pass-rate gains over \emph{No Memory} by chain position in MemoryArena shopping and search. $N$ below the axis counts the chains reaching each position.}
\label{fig:depth}
\end{figure}

\paragraph{Effect of chain depth.}
Figure~\ref{fig:depth} shows how the gain over the \emph{No Memory} baseline varies with chain position in shopping and search. Every shopping chain has six purchases, while search chains run longer. \emph{\leto{}}'s gain grows with position in both. In shopping it rises from 1.3 percentage points at the first purchase to 7.3 points at the sixth. In search it rises from zero at the first question to approximately 14 points at the eighth, where every other system is below 10 points and most remain close to zero. A later step depends on what earlier rounds established. \emph{\leto{}} serves it as the task state in every later round, so its contribution grows as the chain proceeds. \emph{Letta} holds the larger gain in shopping, where a later purchase depends on only a few earlier ones. In search \emph{Letta} reaches 7.2 points at the eighth question, half of \emph{\leto{}}'s gain.

\subsection{Scenario 3: Long-Context Understanding}
\label{sec:exp-longcontext}

Long-context understanding requires an agent to recover details and relationships from earlier conversations. Questions often identify the information sought more explicitly than in the other scenarios.

\paragraph{Main results.}
Table~\ref{tab:longcontext} reports accuracy on LoCoMo and LongMemEval, with the tokens each system retrieves per question and its evidence recall. \emph{\leto{}} achieves the highest accuracy among the memory systems on both benchmarks, 71.82\% on LoCoMo and 73.20\% on LongMemEval. It is 2.79 percentage points lower than \emph{RAG} on LoCoMo and 6.80 points higher on LongMemEval. \emph{RAG} retrieves about $1.7\times$ and eight times as many tokens. Retrieval over the raw history is already effective because these questions identify the information sought, which suggests that the benefit of memory organization varies with how explicitly the request identifies the relevant history. Results by question category are in Appendix~\ref{app:longcontext}.

\begin{table}[t]
  \centering
  \footnotesize
  \setlength{\tabcolsep}{2pt}
  \begin{tabular}{l|ccc|ccc}
    \toprule
    \multicolumn{1}{c|}{} & \multicolumn{3}{c|}{\textbf{LoCoMo}} & \multicolumn{3}{c}{\textbf{LongMemEval}} \\
    \cmidrule[\heavyrulewidth](lr){2-4} \cmidrule[\heavyrulewidth](lr){5-7}
    \multicolumn{1}{l|}{\textbf{System}} & \multicolumn{1}{c}{\textbf{Tokens}} & \multicolumn{1}{c}{\textbf{Recall}} & \multicolumn{1}{c|}{\textbf{Acc.}} & \multicolumn{1}{c}{\textbf{Tokens}} & \multicolumn{1}{c}{\textbf{Recall}} & \multicolumn{1}{c}{\textbf{Acc.}} \\
    \midrule[\heavyrulewidth]
    \multicolumn{7}{l}{\textbf{\emph{No History}}} \\
    \midrule
    No Memory & -- & -- & 4.55 & -- & -- & 6.40 \\
    \midrule
    \multicolumn{7}{l}{\textbf{\emph{Full History}}} \\
    \midrule
    Standalone & 27{,}834 & -- & 74.42 & 111{,}569 & -- & 51.60 \\
    RAG & 1{,}999 & -- & 74.61 & 9{,}813 & -- & 66.40 \\
    \midrule
    \multicolumn{7}{l}{\textbf{\emph{Compared Memory Systems}}} \\
    \midrule
    Mem0 & 745 & 56.69 & 67.86 & 1{,}113 & 79.18 & 69.20 \\
    LangMem & 1{,}757 & 43.07 & 32.08 & 1{,}735 & 67.80 & 37.60 \\
    Hindsight & 608 & 46.51 & 65.13 & 701 & 70.61 & 70.00 \\
    Letta & 55 & 26.24 & 51.36 & 86 & 29.59 & 36.60 \\
    Zep & 479 & 29.79 & 41.36 & 2{,}039 & 69.36 & 48.00 \\
    Memobase & 1{,}027 & 24.79 & 13.18 & 1{,}025 & 34.40 & 15.00 \\
    \midrule
    \multicolumn{7}{l}{\textbf{\emph{Our System}}} \\
    \midrule
    \rowcolor{oursbg} \leto{} & 1{,}192 & \textbf{86.58} & \textbf{71.82} & 1{,}185 & \textbf{95.38} & \textbf{73.20} \\
    \bottomrule
  \end{tabular}
  \caption{Long-context understanding on LoCoMo and LongMemEval\textsubscript{S}. Tokens: average tokens of retrieved memory per question (the full history for \emph{Standalone}, which uses gpt-4o-mini). Recall: evidence recall (\%), measured for the memory systems. Acc.: accuracy (\%). Bold: best value among the memory systems.}
  \label{tab:longcontext}
\end{table}

\paragraph{Evidence coverage.}
Evidence recall is the fraction of an annotated evidence message's content words that appear in the served context, a lexical proxy for whether the supporting evidence reaches the agent (Appendix~\ref{app:longcontext}). \emph{\leto{}} has the highest recall on both benchmarks, 86.58\% on LoCoMo and 95.38\% on LongMemEval. The other memory systems reach at most 56.69\% and 79.18\%. Serving the source of the retrieved facts therefore keeps most of the evidence in a small context. The remaining difference from \emph{RAG} on LoCoMo is concentrated in single-hop questions. Approximately 40\% of those answered correctly by \emph{RAG} but not by \emph{\leto{}} have an evidence message that no extracted fact points to, so access through the fact index still depends on extraction coverage.

\section{Conclusion}
\label{sec:conclusion}

The value of agent memory lies in the responses it enables, and fidelity to past conversations does not ensure that value. We viewed memory as a prediction problem and presented \emph{\leto{}}, a memory system that decides what to memorize and what to serve by the utility of memory to future responses. \emph{\leto{}} leads the compared memory systems in three scenarios that place different demands on memory, which suggests that utility is a general criterion for memory design. We hope this view moves memory systems from how faithfully they record the past toward how much they improve what the agent does next. As future work, we will learn utility forecasts from feedback so that memory systems better anticipate the needs of future interactions. We will also study how to attribute an outcome to an individual memory and how utility transfers across agents.

\section*{Limitations}

\emph{\leto{}} trades storage and write-time computation for the ability to recover what extraction did not predict would matter. Keeping every conversation beneath the fact index removes the loss of discarded detail, but it does not make every detail reachable: a message that no fact points to is served only when another fact's source contains it. Extraction coverage therefore remains the bound on how far the memorization gap can be closed. Serving standing information before the request bridges the serving gap where a request leaves a standing constraint unstated, at the cost of context that a given request may not need. The profile and task state are generated summaries that can misstate what the conversations established. The utility index is updated from outcomes, but an outcome is shared by every memory served for that response. The signal is therefore coarse: an unnecessary memory can be rewarded and a correct one penalized. A memory that is never served never earns an observation. More generally, utility is a property of the memory system together with the agent it serves. The same served memory can help one agent and not another, so what \emph{\leto{}} learns about a memory's utility is specific to the agent and the tasks it was observed on. Finally, the approach addresses what to memorize and what to serve but not how the agent uses what it is served; an agent can still overlook useful served memory~\citep{liu2024lost}.

\bibliography{references}
\appendix
\section{Benchmarks}
\label{app:benchmarks}

Table~\ref{tab:benchmarks} lists the five benchmarks with their number of graded units and their memory-store scope, along with the metrics reported.

\begin{table*}[t]
  \centering
  \footnotesize
  \begin{tabular}{llrll}
    \toprule
    \textbf{Scenario} & \textbf{Benchmark} & \textbf{\# Samples} & \textbf{Store} & \textbf{Metrics} \\
    \midrule[\heavyrulewidth]
    \multirow{2}{*}{Personalization} & PERMA C-S / C-M & 423 / 157 & One user & MCQ Accuracy, Comp. \\
     & PrefEval & 1{,}000 & One item & Accuracy \\
    \midrule
    Long-horizon tasks & MemoryArena & 4{,}850$^{a}$ & One chain & PS, SR \\
    \midrule
    \multirow{2}{*}{Long-context} & LoCoMo & 1{,}540 & One conversation & Accuracy \\
     & LongMemEval\textsubscript{S} & 500 & One question & Accuracy \\
    \bottomrule
  \end{tabular}
  \par\smallskip
  \parbox{\linewidth}{\footnotesize
  $^{a}$ Travel 1{,}869 traveler-rounds; shopping 900 purchase steps; search 1{,}641 questions; math 354
  and physics 86 subtasks.}
  \caption{Benchmarks and memory-store scope. Samples count graded units. C-S / C-M: PERMA's clean single- / multi-domain conditions. Metrics: multiple-choice question (MCQ) accuracy; task completion (Comp.); progress score (PS); success rate (SR).}
  \label{tab:benchmarks}
\end{table*}

\section{Useful and Harmful Rates}
\label{app:rates}

The difference between a system's level and \emph{No Memory}'s on a benchmark is the balance of two outcomes on the same graded units. A unit that passes with memory and fails under \emph{No Memory} has positive utility in Equation~\ref{eq:utility-def} and counts as \emph{useful}; the reverse is \emph{harmful}; a unit with the same result in both runs is unchanged. The useful rate minus the harmful rate is the difference in level. Table~\ref{tab:useful} gives the useful rate and Table~\ref{tab:harmful} the harmful rate on every benchmark. These are paired observed outcomes, not attribution to an individual memory. \emph{No Memory} passes few units on PrefEval and on LoCoMo and LongMemEval, so every system's harmful rate there is small and the useful rate alone orders the systems. \emph{\leto{}}'s useful rate exceeds its harmful rate on every evaluated outcome. Among settings with enough \emph{No Memory} successes to assess degradation, \emph{\leto{}} has the lowest harmful rate among the memory systems on PERMA MCQ and C-M completion and on MemoryArena shopping and math. On PERMA MCQ, \emph{Standalone}'s harmful rate is approximately twice \emph{\leto{}}'s.

\begin{table*}[t]
  \centering
  \footnotesize
  \setlength{\tabcolsep}{2pt}
  \begin{tabular}{ll|cc|cccccc|>{\columncolor{oursbg}}c}
    \toprule
    \multicolumn{2}{c|}{} & \multicolumn{2}{c|}{\textbf{\emph{Full History}}} & \multicolumn{6}{c|}{\textbf{\emph{Compared Memory Systems}}} & \multicolumn{1}{c}{\textbf{\emph{Our System}}} \\
    \cmidrule[\heavyrulewidth](lr){3-4} \cmidrule[\heavyrulewidth](lr){5-10} \cmidrule[\heavyrulewidth](lr){11-11}
    \multicolumn{1}{l}{\textbf{Benchmark}} & \multicolumn{1}{l|}{\textbf{Outcome}} & \multicolumn{1}{c}{\textbf{Standalone}} & \multicolumn{1}{c|}{\textbf{RAG}} & \multicolumn{1}{c}{\textbf{Mem0}} & \multicolumn{1}{c}{\textbf{LangMem}} & \multicolumn{1}{c}{\textbf{Hindsight}} & \multicolumn{1}{c}{\textbf{Letta}} & \multicolumn{1}{c}{\textbf{Zep}} & \multicolumn{1}{c|}{\textbf{Memobase}} & \multicolumn{1}{c}{\textbf{\leto{}}} \\
    \midrule[\heavyrulewidth]
    \multirow{4}{*}{PERMA} & C-S MCQ & 27.66 & 19.39 & 19.15 & 23.17 & 19.15 & 13.48 & 18.68 & 22.70 & \textbf{27.42} \\
     & C-S Comp. & 13.95 & 15.13 & 12.77 & 13.00 & 12.77 & 10.87 & 12.29 & \textbf{13.48} & 12.53 \\
     & C-M MCQ & 21.66 & 17.83 & 15.29 & 16.56 & 14.65 & 12.74 & 9.55 & 16.56 & \textbf{19.11} \\
     & C-M Comp. & 14.01 & 15.29 & 14.65 & 12.10 & 14.01 & 11.46 & 14.01 & \textbf{18.47} & \textbf{18.47} \\
    \midrule
    PrefEval & Accuracy & 2.50 & 12.50 & 32.70 & 39.90 & 43.00 & 17.20 & 8.80 & 27.20 & \textbf{59.30} \\
    \midrule
    \multirow{5}{*}{MemoryArena} & Shopping & 17.89 & 13.33 & 7.78 & 15.00 & 7.33 & \textbf{20.44} & 12.56 & 12.89 & 15.33 \\
     & Travel & 2.30 & 1.50 & 0.00 & 0.00 & 0.00 & 0.00 & 0.00 & 0.00 & \textbf{2.73} \\
     & Search & 2.44 & 4.14 & 2.80 & 3.47 & 2.38 & 3.47 & 4.75 & 3.35 & \textbf{5.91} \\
     & Math & 10.45 & 12.15 & 9.32 & 3.95 & 8.19 & 9.89 & 7.06 & 4.80 & \textbf{10.17} \\
     & Physics & 18.60 & 15.12 & 17.44 & 6.98 & 12.79 & \textbf{23.26} & 6.98 & 10.47 & 19.77 \\
    \midrule
    LoCoMo & Accuracy & 70.45 & 70.58 & 64.09 & 29.42 & 61.75 & 47.53 & 38.38 & 10.65 & \textbf{67.86} \\
    \midrule
    LongMemEval & Accuracy & 47.20 & 61.20 & 64.40 & 32.40 & 65.40 & 31.60 & 43.20 & 9.20 & \textbf{69.00} \\
    \bottomrule
  \end{tabular}
  \caption{Useful rates against \emph{No Memory}: the share of each benchmark's graded units that the system passes and \emph{No Memory} fails (\%). Shopping uses step success. \emph{Standalone} places the full history in the agent's context (gpt-4o-mini; gpt-5-mini on MemoryArena). C-S / C-M: clean single- / multi-domain; Comp.: completion; Bold: highest value per row among the memory systems.}
  \label{tab:useful}
\end{table*}

\begin{table*}[t]
  \centering
  \footnotesize
  \setlength{\tabcolsep}{2pt}
  \begin{tabular}{ll|cc|cccccc|>{\columncolor{oursbg}}c}
    \toprule
    \multicolumn{2}{c|}{} & \multicolumn{2}{c|}{\textbf{\emph{Full History}}} & \multicolumn{6}{c|}{\textbf{\emph{Compared Memory Systems}}} & \multicolumn{1}{c}{\textbf{\emph{Our System}}} \\
    \cmidrule[\heavyrulewidth](lr){3-4} \cmidrule[\heavyrulewidth](lr){5-10} \cmidrule[\heavyrulewidth](lr){11-11}
    \multicolumn{1}{l}{\textbf{Benchmark}} & \multicolumn{1}{l|}{\textbf{Outcome}} & \multicolumn{1}{c}{\textbf{Standalone}} & \multicolumn{1}{c|}{\textbf{RAG}} & \multicolumn{1}{c}{\textbf{Mem0}} & \multicolumn{1}{c}{\textbf{LangMem}} & \multicolumn{1}{c}{\textbf{Hindsight}} & \multicolumn{1}{c}{\textbf{Letta}} & \multicolumn{1}{c}{\textbf{Zep}} & \multicolumn{1}{c|}{\textbf{Memobase}} & \multicolumn{1}{c}{\textbf{\leto{}}} \\
    \midrule[\heavyrulewidth]
    \multirow{4}{*}{PERMA} & C-S MCQ & 3.31 & 2.60 & 2.84 & 2.13 & 3.78 & 10.64 & 4.02 & 3.07 & \textbf{1.65} \\
     & C-S Comp. & 6.62 & 5.67 & 6.62 & \textbf{5.67} & 6.86 & 10.64 & 8.04 & 5.91 & 6.38 \\
     & C-M MCQ & 7.01 & 6.37 & 4.46 & 5.73 & 5.10 & 14.65 & 7.64 & 5.73 & \textbf{3.82} \\
     & C-M Comp. & 10.83 & 14.01 & 12.74 & 10.19 & 14.65 & 17.20 & 10.83 & 14.01 & \textbf{6.37} \\
    \midrule
    PrefEval & Accuracy & 0.60 & 0.80 & 0.60 & 0.60 & 0.20 & \textbf{0.00} & 0.80 & 0.40 & 0.20 \\
    \midrule
    \multirow{5}{*}{MemoryArena} & Shopping & 9.67 & 9.00 & 11.56 & 10.33 & 9.11 & 8.78 & 10.33 & 9.89 & \textbf{8.00} \\
     & Travel & 0.00 & 0.00 & 0.00 & 0.00 & 0.00 & 0.00 & 0.00 & 0.00 & 0.00 \\
     & Search & 2.38 & 1.58 & 1.28 & 1.58 & 1.34 & 1.58 & \textbf{1.16} & 1.40 & 1.46 \\
     & Math & 3.39 & 3.11 & 4.24 & 3.67 & 2.82 & 2.82 & 4.52 & 4.52 & \textbf{2.26} \\
     & Physics & 3.49 & 3.49 & \textbf{2.33} & 9.30 & 4.65 & 3.49 & 5.81 & 6.98 & 3.49 \\
    \midrule
    LoCoMo & Accuracy & 0.58 & 0.52 & 0.78 & 1.88 & 1.17 & 0.71 & 1.56 & 2.01 & \textbf{0.58} \\
    \midrule
    LongMemEval & Accuracy & 2.00 & 1.20 & 1.60 & 1.20 & 1.80 & 1.40 & 1.60 & \textbf{0.60} & 2.20 \\
    \bottomrule
  \end{tabular}
  \caption{Harmful rates against \emph{No Memory}: the share of each benchmark's graded units that \emph{No Memory} passes and the system fails (\%). Shopping uses step success. \emph{Standalone} places the full history in the agent's context (gpt-4o-mini; gpt-5-mini on MemoryArena). C-S / C-M: clean single- / multi-domain; Comp.: completion; Bold: lowest value per row among the memory systems.}
  \label{tab:harmful}
\end{table*}

\section{Long-Context Results by Category}
\label{app:longcontext}

Tables~\ref{tab:locomo} and~\ref{tab:longmemeval} give the long-context results by question category. LoCoMo covers single-hop and multi-hop as well as temporal and open-domain questions; its adversarial category of questions with no answer in the conversation is excluded as in \emph{Mem0}'s evaluation~\citep{chhikara2025mem0}. The LoCoMo difference from \emph{RAG} is concentrated in single-hop questions, where \emph{RAG} reaches 82.05\% against \emph{\leto{}}'s 76.69\%. The largest LongMemEval advantage over \emph{Hindsight} is in temporal reasoning, where \emph{\leto{}} achieves 73.68\% against 62.41\%.

\begin{table*}[t]
  \centering
  \footnotesize
  \begin{tabular}{l|cccc|c|cc}
    \toprule
    \multicolumn{1}{c|}{} & \multicolumn{4}{c|}{\textbf{Category}} & \multicolumn{1}{c|}{} & \multicolumn{2}{c}{\textbf{Overall}} \\
    \cmidrule[\heavyrulewidth](lr){2-5} \cmidrule[\heavyrulewidth](lr){7-8}
    \multicolumn{1}{l|}{\textbf{System}} & \multicolumn{1}{c}{\textbf{Single-hop}} & \multicolumn{1}{c}{\textbf{Multi-hop}} & \multicolumn{1}{c}{\textbf{Temporal}} & \multicolumn{1}{c|}{\textbf{Open-domain}} & \multicolumn{1}{c|}{\textbf{Tokens}} & \multicolumn{1}{c}{\textbf{Accuracy}} & \multicolumn{1}{c}{\textbf{F1}} \\
    \midrule[\heavyrulewidth]
    \multicolumn{8}{l}{\textbf{\emph{No History}}} \\
    \midrule
    No Memory & 5.59 & 2.48 & 1.56 & 11.46 & -- & 4.55 & 2.99 \\
    \midrule
    \multicolumn{8}{l}{\textbf{\emph{Full History}}} \\
    \midrule
    Standalone (gpt-4o-mini) & 86.33 & 72.70 & 48.29 & 62.50 & 27{,}834 & 74.42 & 46.52 \\
    RAG & 82.05 & 62.41 & 72.27 & 53.12 & 1{,}999 & 74.61 & 54.32 \\
    \midrule
    \multicolumn{8}{l}{\textbf{\emph{Compared Memory Systems}}} \\
    \midrule
    Mem0 & 75.98 & 58.87 & 60.44 & 47.92 & 745 & 67.86 & 49.90 \\
    LangMem & 39.71 & 35.46 & 6.54 & 40.62 & 1{,}757 & 32.08 & 22.61 \\
    Hindsight & 67.66 & \textbf{65.96} & 62.62 & 48.96 & 608 & 65.13 & 48.72 \\
    Letta & 63.85 & 50.35 & 24.92 & 33.33 & 55 & 51.36 & 39.74 \\
    Zep & 41.74 & 35.11 & 45.17 & 43.75 & 479 & 41.36 & 31.43 \\
    Memobase & 14.98 & 8.51 & 5.61 & 36.46 & 1{,}027 & 13.18 & 10.01 \\
    \midrule
    \multicolumn{8}{l}{\textbf{\emph{Our System}}} \\
    \midrule
    \rowcolor{oursbg} \leto{} & \textbf{76.69} & 61.70 & \textbf{72.59} & \textbf{56.25} & 1{,}192 & \textbf{71.82} & \textbf{51.86} \\
    \bottomrule
  \end{tabular}
  \caption{LoCoMo results (\%). Tokens: average tokens of retrieved memory per question (the full history for \emph{Standalone}). Bold: best value among the memory systems.}
  \label{tab:locomo}
\end{table*}

\begin{table*}[t]
  \centering
  \footnotesize
  \setlength{\tabcolsep}{2.5pt}
  \begin{tabular}{l|ccccccc|c|c}
    \toprule
    \multicolumn{1}{c|}{} & \multicolumn{7}{c|}{\textbf{Category}} & \multicolumn{1}{c|}{} & \multicolumn{1}{c}{\textbf{Overall}} \\
    \cmidrule[\heavyrulewidth](lr){2-8} \cmidrule[\heavyrulewidth](lr){10-10}
    \multicolumn{1}{l|}{\textbf{System}} & \multicolumn{1}{c}{\textbf{\makecell{SS\\user}}} & \multicolumn{1}{c}{\textbf{\makecell{SS\\assistant}}} & \multicolumn{1}{c}{\textbf{\makecell{SS\\preference}}} & \multicolumn{1}{c}{\textbf{\makecell{Multi-\\session}}} & \multicolumn{1}{c}{\textbf{\makecell{Knowledge\\update}}} & \multicolumn{1}{c}{\textbf{\makecell{Temporal\\reasoning}}} & \multicolumn{1}{c|}{\textbf{Abstention}} & \multicolumn{1}{c|}{\textbf{Tokens}} & \multicolumn{1}{c}{\textbf{Accuracy}} \\
    \midrule[\heavyrulewidth]
    \multicolumn{10}{l}{\textbf{\emph{No History}}} \\
    \midrule
    No Memory & 8.57 & 3.57 & 0.00 & 9.02 & 7.69 & 4.51 & 100.00 & -- & 6.40 \\
    \midrule
    \multicolumn{10}{l}{\textbf{\emph{Full History}}} \\
    \midrule
    Standalone (gpt-4o-mini) & 65.71 & 85.71 & 13.33 & 38.35 & 66.67 & 42.86 & 66.67 & 111{,}569 & 51.60 \\
    RAG & 90.00 & 100.00 & 20.00 & 56.39 & 69.23 & 58.65 & 80.00 & 9{,}813 & 66.40 \\
    \midrule
    \multicolumn{10}{l}{\textbf{\emph{Compared Memory Systems}}} \\
    \midrule
    Mem0 & \textbf{94.29} & \textbf{91.07} & 50.00 & 59.40 & \textbf{82.05} & 53.38 & 73.33 & 1{,}113 & 69.20 \\
    LangMem & 54.29 & 30.36 & 50.00 & 22.56 & 60.26 & 30.83 & 83.33 & 1{,}735 & 37.60 \\
    Hindsight & \textbf{94.29} & 64.29 & \textbf{53.33} & \textbf{67.67} & 75.64 & 62.41 & 70.00 & 701 & 70.00 \\
    Letta & 51.43 & 53.57 & 33.33 & 24.81 & 51.28 & 25.56 & 76.67 & 86 & 36.60 \\
    Zep & 67.14 & 55.36 & 46.67 & 36.09 & 60.26 & 39.85 & 73.33 & 2{,}039 & 48.00 \\
    Memobase & 18.57 & 7.14 & 20.00 & 8.27 & 34.62 & 10.53 & \textbf{93.33} & 1{,}025 & 15.00 \\
    \midrule
    \multicolumn{10}{l}{\textbf{\emph{Our System}}} \\
    \midrule
    \rowcolor{oursbg} \leto{} & 90.00 & 89.29 & 46.67 & 61.65 & 75.64 & \textbf{73.68} & 63.33 & 1{,}185 & \textbf{73.20} \\
    \bottomrule
  \end{tabular}
  \caption{LongMemEval\textsubscript{S} results (\%). The 30 abstention questions also count within their question types. SS: single-session. Tokens: average tokens of retrieved memory per question (the full history for \emph{Standalone}). Bold: best value among the memory systems.}
  \label{tab:longmemeval}
\end{table*}

\begin{figure*}[t]
  \centering
  \includegraphics[width=.92\linewidth]{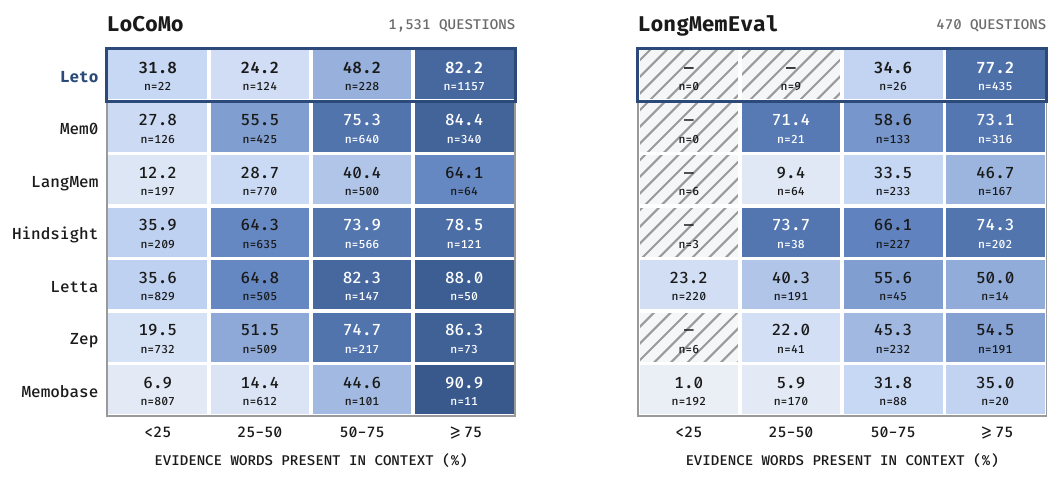}
  \caption{Answer accuracy by evidence-recall band. Each cell shows its number of questions; hatched cells contain fewer than ten.}
  \label{fig:recall}
\end{figure*}

\paragraph{Evidence recall.}
Evidence recall is the fraction of an evidence message's content words appearing in the served context, averaged first over evidence messages and then over questions. Content words are lowercased words with at least four letters, excluding stop words. We use annotated evidence for 1{,}531 LoCoMo questions and the best-matching turn from each annotated answer session for 470 answerable LongMemEval questions. Figure~\ref{fig:recall} groups questions by evidence recall. On LoCoMo, every memory system achieves higher accuracy in the highest recall band than in the lowest. On LongMemEval, context from the annotated evidence session reaches the agent for 98\% of questions under \emph{\leto{}} and 99\% under \emph{RAG}. Accuracy among these questions is 75\% and 67\%, respectively.

\end{document}